\documentclass[10pt,twocolumn,letterpaper]{article}

\usepackage{cvpr}
\usepackage{times}
\usepackage{epsfig}
\usepackage{graphicx}
\usepackage{amsmath}
\usepackage{amssymb}
\usepackage{subfig}
\usepackage{color}
\usepackage{algorithm}
\usepackage{algorithmic}
\usepackage{bm}
\usepackage{multirow}  
\usepackage{booktabs}
\usepackage[table]{xcolor}
\definecolor{mygray}{gray}{.92}

\makeatletter
\newcommand{\thickhline}{%
    \noalign {\ifnum 0=`}\fi \hrule height 1pt
    \futurelet \reserved@a \@xhline
}

\newcommand{\blue}[1]{{\color{blue}{#1}}}
\newcommand{\red}[1]{{\color{red}{#1}}}
\usepackage[pagebackref=true,breaklinks=true,letterpaper=true,colorlinks,bookmarks=false,citecolor=green]{hyperref}

\cvprfinalcopy % *** Uncomment this line for the final submission

\def\cvprPaperID{680} % *** Enter the CVPR Paper ID here

\ifcvprfinal\fi
\begin{document}

%%%%%%%%% TITLE
\title{Learning Multi-Granular Hypergraphs for Video-Based Person Re-Identification}

\author{Yichao Yan$^{1}$\thanks{indicates equal contributions; ${}^{\dagger}$ indicates corresponding author} ,
Jie Qin$^{1}$\footnotemark[1]\, ${}^{\dagger}$,
Jiaxin Chen$^{1}$,
Li Liu$^{1}$,
Fan Zhu$^{1}$,
Ying Tai$^{2}$,
and Ling Shao$^{1}$
\\
\noindent
$^{1}$
Inception Institute of Artificial Intelligence (IIAI), Abu Dhabi, UAE \qquad \\
$^{2}$ Tencent YouTu Lab, Shanghai, China \qquad 
\\
$^{}$ {\tt\small \{firstname.lastname\}@inceptioniai.org,} \tt\small yingtai@tencent.com\\
}

\maketitle
\thispagestyle{empty}

%%%%%%%%% ABSTRACT
\begin{abstract}
   Video-based person re-identification (re-ID) is an important research topic in computer vision. The key to tackling the challenging task is to exploit both spatial and temporal clues in video sequences. %in order to learn robust representations. 
   In this work, we propose a novel graph-based framework, namely \textbf{Multi-Granular Hypergraph} (MGH), to pursue better representational capabilities by modeling spatiotemporal dependencies in terms of multiple granularities. Specifically, hypergraphs with different spatial granularities are constructed using various levels of part-based features across the video sequence. In each hypergraph, different temporal granularities are captured by hyperedges that connect a set of graph nodes (i.e., part-based features) across different temporal ranges. Two critical issues (misalignment and occlusion) are explicitly addressed by the proposed hypergraph propagation and feature aggregation schemes. Finally, we further enhance the overall video representation by learning more diversified graph-level representations of multiple granularities based on mutual information minimization. Extensive experiments on three widely-adopted benchmarks clearly demonstrate the effectiveness of the proposed framework. Notably, \textbf{90.0\%} top-1 accuracy on \textbf{MARS} is achieved using MGH, outperforming the state-of-the-arts. Code is available at {\tt\small \url{https://github.com/daodaofr/hypergraph_reid}}.
   %we exploit spatial clues by dividing human feature into multi-level partitions to represent human appearance. We leverage temporal clues by building a hypergraph based on part-level features. Human parts in a sequence are connected with different hyperedges to exploit short-term and long-term correlations. Part-level features are updated via message propagation within the graph, and discriminative sequence-level is generated via graph pooling. Extensive experiments on three widely-adopted video-based person re-ID datasets have well demonstrated the effectiveness of the proposed framework. Notably, for the first time, 90.0\% rank-1 accuracy on the MARS dataset is achieved using the proposed model, outperforming the existing state-of-the-arts.
\end{abstract}

%%%%%%%%% BODY TEXT
\section{Introduction}
\begin{figure}[t]
\setlength{\abovecaptionskip}{-0.2mm}
 \centering
 \includegraphics[width=\linewidth]{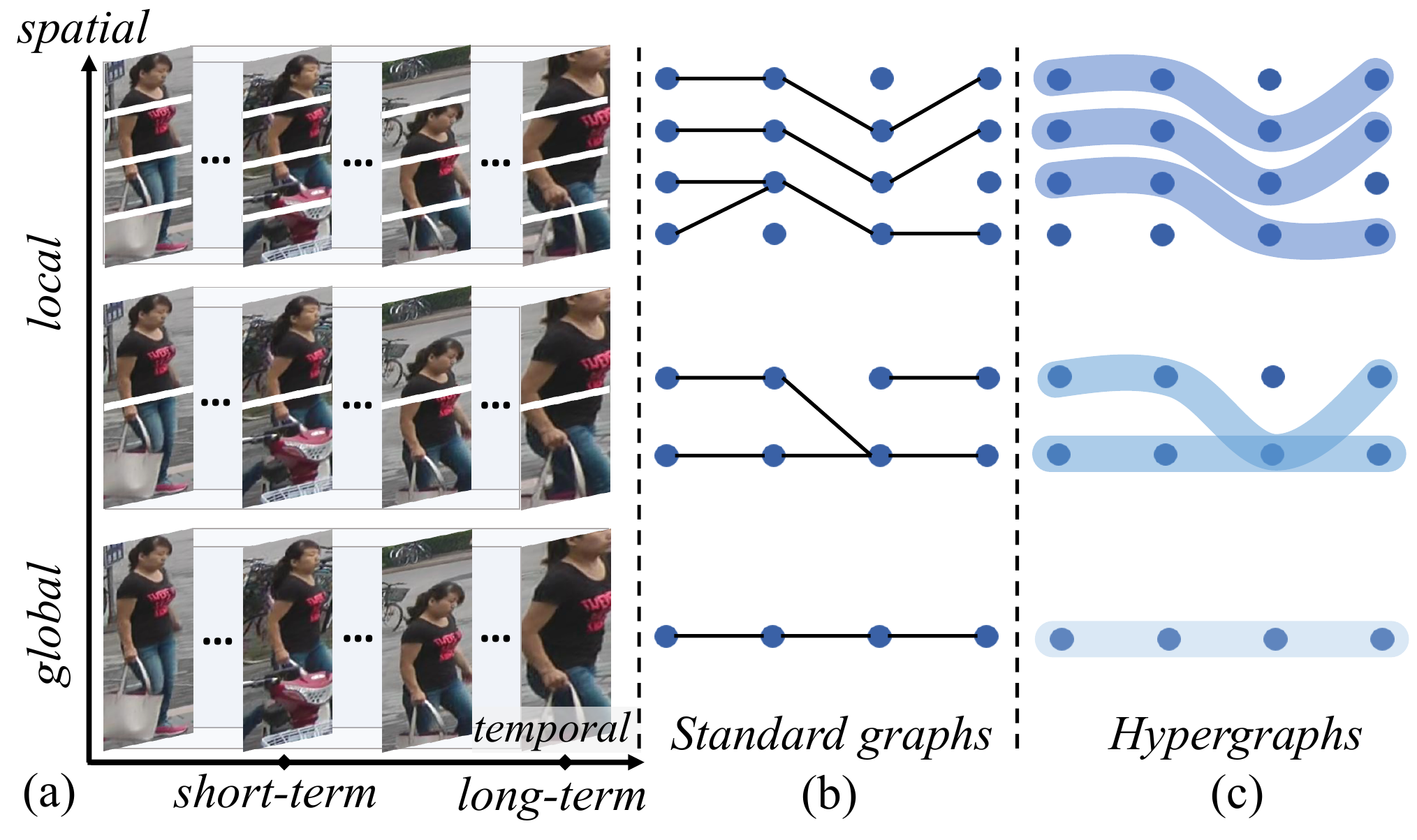}
 \caption{(a) Illustration of multi-granular spatial and temporal clues, which provide important insights into addressing the challenges of misalignment and occlusion in video-based person re-ID. (b) Standard graphs can only model the dependency between node pairs, lacking the capability of modeling long-term temporal dependency. (c) Hypergraphs can model both short-term and long-term dependencies by associating multiple nodes within a single hyperedge.}
 \label{fig:intro}
 \vspace{-4mm}
\end{figure}

Person re-identification (re-ID) aims at associating individuals across non-overlapping cameras, with great potential in surveillance-related applications. As such, significant efforts have been made in the past few years to address the challenging task. In parallel with the prevalence of image-based person re-ID, person re-ID based on video sequences has also recently emerged. This is because the richer information in videos can be utilized to reduce visual ambiguities, especially for people sharing similar appearances. The key to solving video-based person re-ID is to concurrently exploit spatial and temporal clues within video sequences. In this sense, this work aims to shed light on two important clues (see Figure~\ref{fig:intro}) for tackling video-based person re-ID.

1) \textbf{Multi-granularity of spatial clues.} As the structural information of the human body is beneficial for person identification, part-based models~\cite{DBLP:conf/eccv/VariorSLXW16,DBLP:conf/cvpr/LiC0H17,DBLP:conf/eccv/SunZYTW18} have generally achieved promising performance in person re-ID. Compared with fixed partitions, multi-granular part-based models~\cite{DBLP:conf/mm/WangYCLZ18,DBLP:conf/cvpr/ZhengDSJGYHJ19} have further enhanced the performance by dividing the human body into multiple granularities. In video-based re-ID, the multi-granular spatial clues are particularly important since different levels of granularities capture discrepancies between different partitions, thus addressing the spatial \emph{misalignment} issue due to inaccurate detection in the video sequence (see Figure~\ref{fig:intro}(a)).  
%\red{The multi-granularity of spatial clues is especially important for video-based re-ID task, as there usually exist different levels of misalignment in the sequence due to inaccurate detection (as shown in Figure~\ref{fig:intro}). Therefore it requires different spatial granularity to capture the discrepancy between local parts.} 
However, in this way, misalignment can only be solved implicitly to some extent. To better tackle spatial misalignment, we need to explicitly align different body parts across the whole sequence in order to achieve more robust re-ID performance.
%However, the misalignment issue can only be solved implicitly to some extent in this way. To better tackle the misalignment issue, we need to explicitly find correlations between different body parts within the whole sequence. If body parts are aligned well across the sequence, more robust re-ID performance could be achieved.
%Existing models typically deal with each part independently \red{and simply perform pooling to aggregate part-based features in the sequence}, neglecting the correlations between different body parts.
It is therefore highly desirable to develop re-ID within a framework which systematically captures correlations between different body partitions, while the same time being able to exploit multiple spatial granularities.

2) \textbf{Multi-granularity of temporal clues.} Temporal clues have been extensively studied by previous video-based re-ID models. Short-term dynamics can be represented by extracting additional optical flow features~\cite{DBLP:conf/iccv/ChungTD17}, while long-term temporal features can be obtained by utilizing 3D CNNs~\cite{DBLP:conf/aaai/LiZH19} or temporal feature aggregators~\cite{DBLP:conf/cvpr/McLaughlinRM16} (\eg, Recurrent Neural Networks, RNNs). However, short- and long-term temporal clues have different functionalities in discriminative feature learning. For example, in Figure~\ref{fig:intro}, there is a partial occlusion w.r.t. the short-term temporal clues; the long-term temporal clues can help reduce its impact. However, only a few works~\cite{DBLP:journals/corr/abs-1908-10049,DBLP:conf/aaai/LiZH19} address this issue. It is thus highly important to design a model that can capture multi-granular temporal clues.

\begin{figure*}[t]
\setlength{\abovecaptionskip}{-0.2mm}
 \centering
 \includegraphics[width=\linewidth]{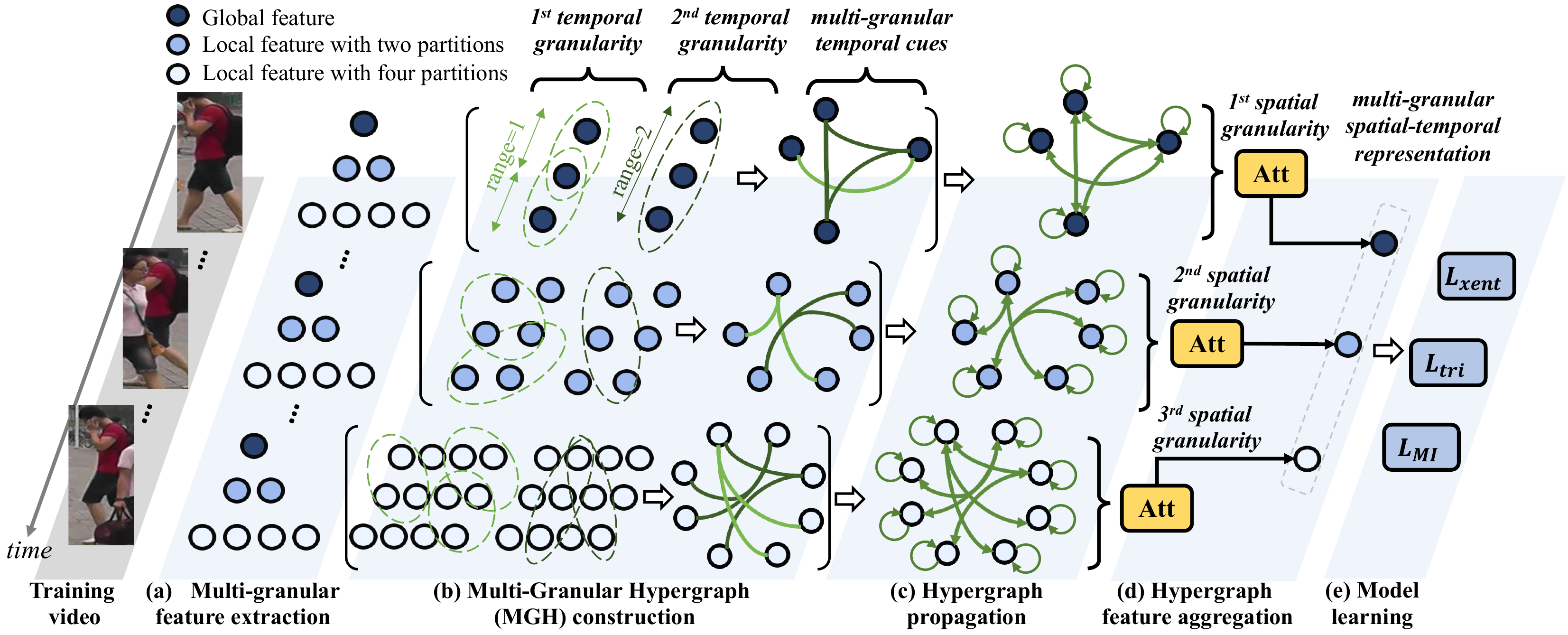}
 \caption{Detailed architecture of the proposed multi-granular hypergraph learning framework for video-based person re-ID. For better visualization, we only illustrate the first three spatial granularities and the first two temporal granularities.}
 \label{fig:method}
 \vspace{-4mm}
\end{figure*}

%we model each video sequence as a set of hypergraphs, where each hypergraph corresponds to a specific spatial granularity and graph nodes refer to global or part-level features. For each hypergraph, graph nodes are connected with each other by a set of hyperedges, where each edge represents a specific temporal granularity. 
To explicitly fulfill the above goal, we propose a novel graph-based framework, named \textbf{Multi-Granular Hypergraph (MGH)}, which simultaneously exploits spatial and temporal clues for video-based person re-ID. As shown in Figure~\ref{fig:method}, we construct a set of hypergraphs to model multiple granularities in a video sequence, with graph nodes representing global or part-level features. Each hypergraph models a specific spatial granularity, while each hyperedge, connecting multiple nodes within a particular temporal range, captures a specific temporal granularity. Node-level features are propagated to form graph-level representations for all hypergraphs, which are aggregated in the final video representation to achieve robust person re-ID.
%The above multi-granular node-level representations are aggregated into a sequence-level feature for robust person representation.
%The correlations between body parts are captured by hyperedges which connect nodes with its nearest neighbors within specific temporal ranges.

Our MGH method has three main advantages. \emph{First}, it seamlessly unifies the learning of spatial and temporal clues into a joint framework, where spatial clues are captured by different hypergraphs, and short- and long-term temporal clues are mined with message propagation through different hyperedges. \emph{Second}, compared with standard graphs which only model correlations between pairs of nodes (see Figure~\ref{fig:intro}(b)), hypergraphs can model high-order dependencies among multiple nodes (see Figure~\ref{fig:intro}(c)). As a result, \emph{misalignment} can be explicitly solved by associating different nodes with their nearest neighbors using hyperedges; meanwhile, \emph{occlusions} can be addressed by modeling multi-granular temporal dependencies with hyperedges across different temporal ranges. \emph{Third}, node-level features can benefit from the spatial and temporal information in the sequence by means of HyperGraph Neural Networks (HGNNs)~\cite{DBLP:conf/aaai/FengYZJG19}, which greatly facilitate information propagation through hyperedges. 
%In this way, video-based person re-ID can be formulated as a hypergraph feature learning task. 
%Thereby, node features can be updated through the propagation of HGNNs for better representation. 
%Based on the above multi-granular representations, an attention module is developed to handle spatial-temporal feature selection to yield more discriminative graph/sequence-level representation.
Our main contributions include:
\begin{itemize}
\setlength{\abovedisplayskip}{0pt}
\setlength{\itemsep}{0pt}
\item We formulate video-based person re-ID as a hypergraph learning task, yielding robust representations based on node propagation and feature aggregation.
\item To capture multi-granular clues, we design a novel HGNN architecture (\ie, MGH) to simultaneously exploit spatial and temporal dependencies in videos.
\item The diversity of graph representations corresponding to different spatial granularities is preserved and enhanced by employing an intuitive loss based on mutual information minimization.
\item MGH achieves promising results on three widely-used re-ID benchmarks. Notably, MGH obtains \textbf{90.0\%} top-1 accuracy on \textbf{MARS}, one of the largest video re-ID datasets, outperforming the state-of-the-art models.
\end{itemize}
%We evaluate the proposed framework on four widely-used video-based person re-ID benchmarks, where the results demonstrate that MGH consistently achieves superior performance compared with current state-of-the-arts.

%-------------------------------------------------------------------------
\section{Related Work}

\textbf{Person Re-identification.}
%Person re-ID has been extensively studied over the past few years. 
Existing works on person re-ID mainly focus on two sub-tasks, \ie, image-based~\cite{DBLP:conf/cvpr/GheissariSH06,DBLP:conf/iccv/ZhengSTWWT15,DBLP:conf/cvpr/ZhengZSCYT17,Chen_2017_CVPR,DBLP:conf/cvpr/BaiBT17} and video-based~\cite{DBLP:conf/cvpr/FarenzenaBPMC10,DBLP:conf/cvpr/ZhengGX12} person re-ID. 
%Video-based re-ID has access to additional temporal information compared with image-based one, and is closer to real scenarios. 
Here, we briefly review some closely related works for video-based re-ID. Early methods tend to employ hand-crafted spatiotemporal features, such as HOG3D~\cite{DBLP:conf/bmvc/KlaserMS08} and SIFT3D~\cite{DBLP:conf/mm/ScovannerAS07}. Other methods try to extract more discriminative descriptors\cite{DBLP:conf/iccv/KaranamLR15,DBLP:conf/iccv/LiuMZH15} or design more effective ranking algorithms~\cite{DBLP:conf/eccv/WangGZW14,DBLP:conf/cvpr/YouWLZ16,DBLP:conf/cvpr/BaiTTL19}.
%Early methods mainly employ hand-crafted features which typically extend 2D features into their spatial-temporal counterparts, such as HOG3D~\cite{DBLP:conf/bmvc/KlaserMS08} and SIFT3D~\cite{DBLP:conf/mm/ScovannerAS07}. Other methods try to extract more discriminative descriptors. Karanam \etal~\cite{DBLP:conf/iccv/KaranamLR15} propose to model a set of frame-wise features with Local Discriminant Analysis. Liu \etal~\cite{DBLP:conf/iccv/LiuMZH15} encode the aligned features within a walking cycle into a spatial-temporal representation to reduce the influence of different poses and viewpoints. Wang \etal~\cite{DBLP:conf/eccv/WangGZW14} propose a video ranking framework by simultaneously selecting the most discriminative fragment in each video and learning a ranking function for re-ID.
Recently, various deep learning models have been proposed and have shown superior performance compared with hand-crafted features. Some works~\cite{DBLP:conf/eccv/ZhengBSWSWT16,DBLP:conf/eccv/LiZG18a,DBLP:conf/cvpr/WuLDYO018,DBLP:conf/eccv/SuhWTML18,DBLP:conf/cvpr/LiB0W18,DBLP:conf/cvpr/ZhangWZ18,DBLP:conf/cvpr/LiuYO17} leverage the powerful learning capability of Convolutional Neural Networks (CNNs) and perform straightforward spatial/temporal pooling on video sequences to generate global representations. However, simply pooling the features may lead to a significant loss of discriminative information. Other methods~\cite{DBLP:conf/cvpr/McLaughlinRM16,DBLP:conf/eccv/YanNSMYY16,DBLP:conf/cvpr/ZhouHWWT17,DBLP:conf/aaai/LiuYZL19,DBLP:conf/cvpr/ChenLXYW18} adopt RNNs and attention mechanisms for more robust temporal feature fusion. However these methods neglect the importance of spatial clues. Another class of methods~\cite{DBLP:conf/iccv/ChungTD17,DBLP:conf/cvpr/McLaughlinRM16} resort to using additional information on optical flow, and adopt a two-stream structure~\cite{DBLP:conf/nips/SimonyanZ14} for discriminative feature learning. However, optical flow only represents local dynamics of adjacent frames, which may introduce noise due to spatial misalignment. 3D CNNs~\cite{DBLP:conf/icml/JiXYY10,DBLP:conf/iccv/TranBFTP15} have also been applied to address video-based person re-ID~\cite{DBLP:conf/aaai/LiZH19}. Despite their promising performance, these networks are computationally expensive and difficult to optimize. In this work, we explicitly explore the multi-granular nature of both spatial and temporal features, yielding more robust representations for video-based re-ID.

\textbf{Hypergraph Learning.}
% Standard graphs can  between graph nodes, which are sub-optimal representations for more complex relationships. To address this issue, 
Graphs are typically leveraged to model relationships between different nodes. Depending on the type of data the nodes represent, graphs have been explored in many computer vision tasks, such as action recognition~\cite{DBLP:conf/eccv/WangG18}, image classification~\cite{DBLP:conf/cvpr/ChenWWG19}, and person re-ID~\cite{DBLP:conf/cvpr/Chen0LSW18,DBLP:conf/eccv/ShenLYCW18}. Recently,
neural networks have been extensively studied for graph learning, leading to the widespread usage of Graph Neural Networks (GNNs)~\cite{DBLP:journals/tnn/ScarselliGTHM09,DBLP:conf/nips/DefferrardBV16,DBLP:conf/nips/DuvenaudMABHAA15,DBLP:conf/miccai/KtenaPFRLGR17,Wang_2019_ICCV}. However, conventional graphs can only model pairwise relationships, which prevents their scalability to data with more complex structures. To this end, hypergraph~\cite{DBLP:conf/nips/ZhouHS06} was introduced to model higher-order relationships between objects of interest, and has been applied to video segmentation~\cite{DBLP:conf/cvpr/HuangLM09} and image retrieval~\cite{DBLP:conf/cvpr/HuangLZM10}. In a similar spirit to GNNs, HyperGraph Neural Networks (HGNNs)~\cite{DBLP:conf/aaai/FengYZJG19,DBLP:journals/corr/abs-1809-02589,DBLP:journals/corr/abs-1901-08150,DBLP:conf/ijcai/JiangWFCG19} have recently been proposed to model correlations in hypergraphs using deep neural networks. Inspired by HGNNs, this work derives a hypergraph that explicitly models spatiotemporal dependency in a video sequence. More importantly, mutiple spatial and temporal granularities are exploited simultaneously in our hypergraph learning framework. As a result, the final global representation exhibits strong discriminability for robust video-based re-ID.

\section{Multi-Granular Hypergraph Learning}
Although deep learning and temporal modeling approaches have greatly improved the performance of video-based person re-ID, it is still difficult to achieve satisfactory results because of occlusion, misalignment, background clutter, and viewpoint changes. To further improve the discriminability of feature representations, this paper aims to explicitly explore the multi-granular nature of spatial and temporal features. To this end, we design a Multi-Granular Hypergraph (MGH) learning framework, which models the high-order correlations between spatial and temporal clues with a hypergraph neural network. The details of the proposed framework are elaborated as follows.

\subsection{Multi-Granular Feature Extraction}
Recent studies~\cite{DBLP:conf/mm/WangYCLZ18,DBLP:conf/cvpr/ZhengDSJGYHJ19} have demonstrated that multi-granular spatial features have the advantage of generating more discriminative representations for human bodies. Inspired by this, we extract multi-granular features for individuals. Specifically, given an image sequence $\mathbf{I} = \{I_1, I_2, ..., I_T\}$ containing $T$ images, we use a backbone CNN model to extract individual feature maps
\begin{equation}\label{eq:1}
    \mathbf{F}_i = {\rm CNN}(I_i), i=1,...,T,
\end{equation}
where $\mathbf{F}_i$ is a 3D tensor with dimensions $C \times H \times W$. $C$ is the channel size, and $H$ and $W$ are the height and width of the feature map, respectively. We then hierarchically divide the feature maps into $p\in\{1,2,4,8\}$ horizontal parts w.r.t. different levels of granularities, and perform average pooling on the divided feature maps to construct a part-level feature vector. For each granularity, the whole sequence generates $N_p=T\times p$ part-level features, which we denote as $\mathbf{h}^{0} = \{\mathbf{h}^{0}_{1},\mathbf{h}^{0}_{2}, ..., \mathbf{h}^{0}_{N_p}\}$.
%$\mathbf{X}_i = \{\mathbf{x}_{i,0},\mathbf{x}_{i,1}, ..., \mathbf{x}_{i,M}\}$, where the $j$-th granularity $\mathbf{x}_{i,j} = \{\mathbf{x}_{i,j,1}, ..., \mathbf{x}_{i,j,2^j}\}$ contains $2^j$ part-level features vectors. 
For example, in each video frame, the first granularity contains a single global vector, while the second and third granularities contain two and four part-level features, respectively, as shown in Figure~\ref{fig:method}(a).

\subsection{Multi-Granular Hypergraph}
After extracting the initial global or part-based features of each individual, \ie initial node features, the next step is to update the node features by learning correlations among different nodes. To generate robust representations, it is necessary to take into account both the spatial and temporal correlations of individual features. Inspired by the recent success of HGNNs~\cite{DBLP:conf/aaai/FengYZJG19,DBLP:conf/ijcai/JiangWFCG19}, we propose a novel hypergraph neural network for spatiotemporal feature learning. To explore the spatial and temporal dependencies within a sequence, HGNNs allow nodes to communicate with their neighbors through message passing within the graph. More importantly, compared to standard graph models, hypergraphs can model the high-order dependency involving multiple nodes, which is more flexible and suitable for modeling the multi-granular correlations in a sequence. 

\textbf{Hypergraph Construction.}
We propose to capture the spatial and temporal dependencies by constructing a set of hypergraphs $\mathcal{G} = \{\mathcal{G}_p\}_{ p\in\{1,2,4,8\}}$, where each hypergraph corresponds to a specific spatial granularity. Concretely, $\mathcal{G}_p = (\mathcal{V}_p, \mathcal{E}_p)$ consists of $N_p$ vertices $\mathcal{V}_p$ and a set of hyperedges $\mathcal{E}_p$. Here, we utilize $v_{i} \in \mathcal{V}_p$, where $i \in \{1,...,N_p\}$ to denote the $i$-th graph node. We define a set of hyperedges to model short- to long-term correlations in the hypergraph. To learn short-term correlations, a hyperedge only connects temporally adjacent features. Mid- and long-range correlations are modeled by hyperedges connecting features of different temporal lengths. Specifically, for each graph node $v_{i}$, we find its $K$ nearest neighbors within specific temporal ranges, according to the feature affinities between nodes. Then we utilize a hyperedge to connect these $K$+1 nodes, as shown in Figure~\ref{fig:method}(b). Mathematically,
\begin{equation}\label{eq:2}
    e_{it} = \{v_{i}, \forall v_{j}\in \mathcal{N}_K(v_{i})\}, s.t. \ |v_i - v_j| \leq T_t,
\end{equation}
where $\mathcal{N}_K$ is the neighborhood set containing the top-$K$ neighbors, $|*|$ denotes the temporal distance between the vertices in the sequence, and $T_t$ is the threshold of temporal range. In our framework, we adopt three thresholds (\ie, $T_1, T_2, T_3$) to model short-term, mid-term and long-term dependencies, respectively. %Therefore, there are a total of $3\times N_p$ hyperedges. 

\begin{figure}
    \centering
    \includegraphics[width=\linewidth]{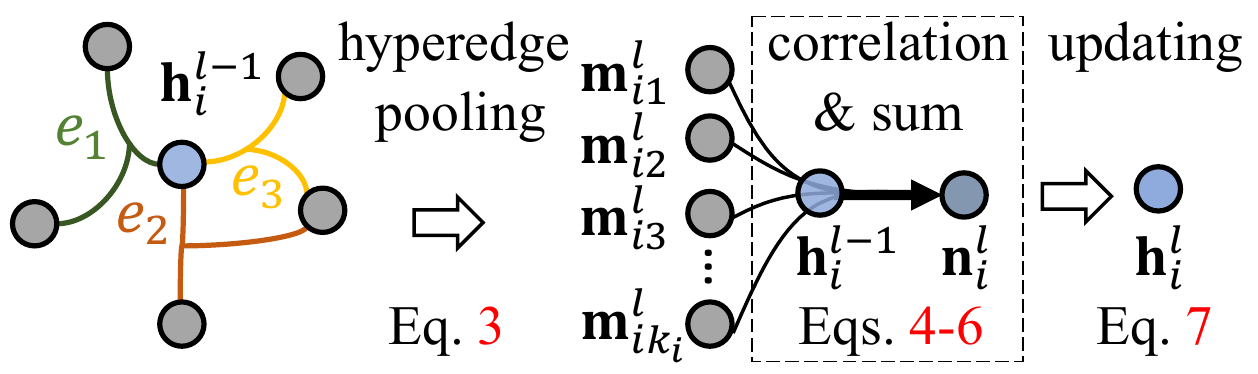}
    \caption{Illustration of node feature propagation.}
    \label{fig:propagation}
    \vspace{-2mm}
\end{figure}

\textbf{Hypergraph Propagation.}
Based on the hypergraphs, we design a hypergraph neural network to propagate graph information and update node features, as illustrated in Figure~\ref{fig:propagation}. Given a node $v_{i}$, we use $Adj(v_{i}) = \{e_1, e_2,..., e_{k_{i}}\}$ to denote all the hyperedges that include this node. These hyperedges contain the nodes that have the highest correlations with $v_{i}$. Then an aggregation operation is defined on the hyperedges to capture feature correlations. Specifically, we average all node features in a hyperedge, except for $v_{i}$, as the hyperedge feature w.r.t. this node:
\begin{equation}\label{eq:3}
    \mathbf{m}^{l}_{ik} = \sum_{j \neq i}^{v_{j} \in e_{k}} \mathbf{h}^{l-1}_{j}, \ \forall e_k \in Adj(v_{i}),
\end{equation}
where $\mathbf{h}^{l-1}_{j}$ denotes the node feature of $v_{j}$ in  layer $l-1$ of the HGNN.
We then calculate the importance of each hyperedge by measuring the correlation between node features and hyperedge features:
\begin{equation}\label{eq:4}
    z_{ik} = \phi(\mathbf{h}^{l-1}_{i}, \mathbf{m}^{l}_{ik}),
\end{equation}
where $\phi$ measures the similarity between features (we employ cosine similarity in our framework). We then utilize the $\operatorname{Softmax}$ function to normalize the importance weights and aggregate the hyperedge messages as follows:
\begin{equation}\label{eq:5}
    \gamma_{ik} =  \frac{{\rm exp}( z_{ik})}{\sum_{j}{\rm exp}(z_{ij})},
\end{equation}
\begin{equation}\label{eq:6}
    \mathbf{n}^{l}_{i} = \sum_{k} \gamma_{ik}\mathbf{m}^{l}_{ik}.
\end{equation}
After obtaining the hypergraph messages, the node features are updated in a fully connected layer by concatenating the previous node features and hyperedge message:
\begin{equation}\label{eq:7}
    \mathbf{h}^l_{i} = \sigma(\mathbf{W}^l [\mathbf{h}^{l-1}_{i}, \mathbf{n}^l_{i}]),
\end{equation}
where $\mathbf{W}^l$ is a weight matrix and $\sigma$ is an activation function.
The above feature updating steps are repeated for $L$ rounds and we obtain a set of output node features $\mathbf{O}_p = \{\mathbf{h}^{L}_{i}\}, \forall {v_{i}\in \mathcal{V}_p}$. We summarize the propagation process of the hypergraph in Algorithm~\ref{alg:1}.

\setlength{\textfloatsep}{10pt}
\begin{algorithm}[t]
\caption{Hypergraph Propagation}
\label{alg:1}
\begin{algorithmic}[1]
\REQUIRE Input sequence $\mathbf{I} = \{I_1, I_2, ..., I_T\}$\\
\ENSURE Hypergraph feature $O_p$ \\
\STATE Extract and pool features by Eq.~\ref{eq:1}: $\mathbf{h}^{0} \leftarrow {I}$ 
\STATE Build the hyperedges by Eq.~\ref{eq:2}
\FOR{$l \leftarrow 1,...,L$}
\STATE Pooling hyperedge features by Eq.~\ref{eq:3}:
 $\mathbf{m}^{l}_{ik} \leftarrow \mathbf{h}^{l-1}$
\STATE Calculate feature correlations and aggregate hyperedge message by Eqs.~\ref{eq:4}-~\ref{eq:6}: $\mathbf{n}^{l}_{i} \leftarrow \mathbf{m}^{l}_{ik}$
\STATE Updating node features by Eq.~\ref{eq:7}: $\mathbf{h}^{l}_{i} \leftarrow \{\mathbf{h}^{l-1}_{i}, \mathbf{n}^{l}_{i} \}$
\ENDFOR
\STATE $\mathbf{O}_p \leftarrow \{\mathbf{h}^{L}_{i}\}$.
\end{algorithmic}
\end{algorithm}

\begin{figure}[t]
\setlength{\abovecaptionskip}{-0.1mm}
    \centering
    \includegraphics[width=\linewidth]{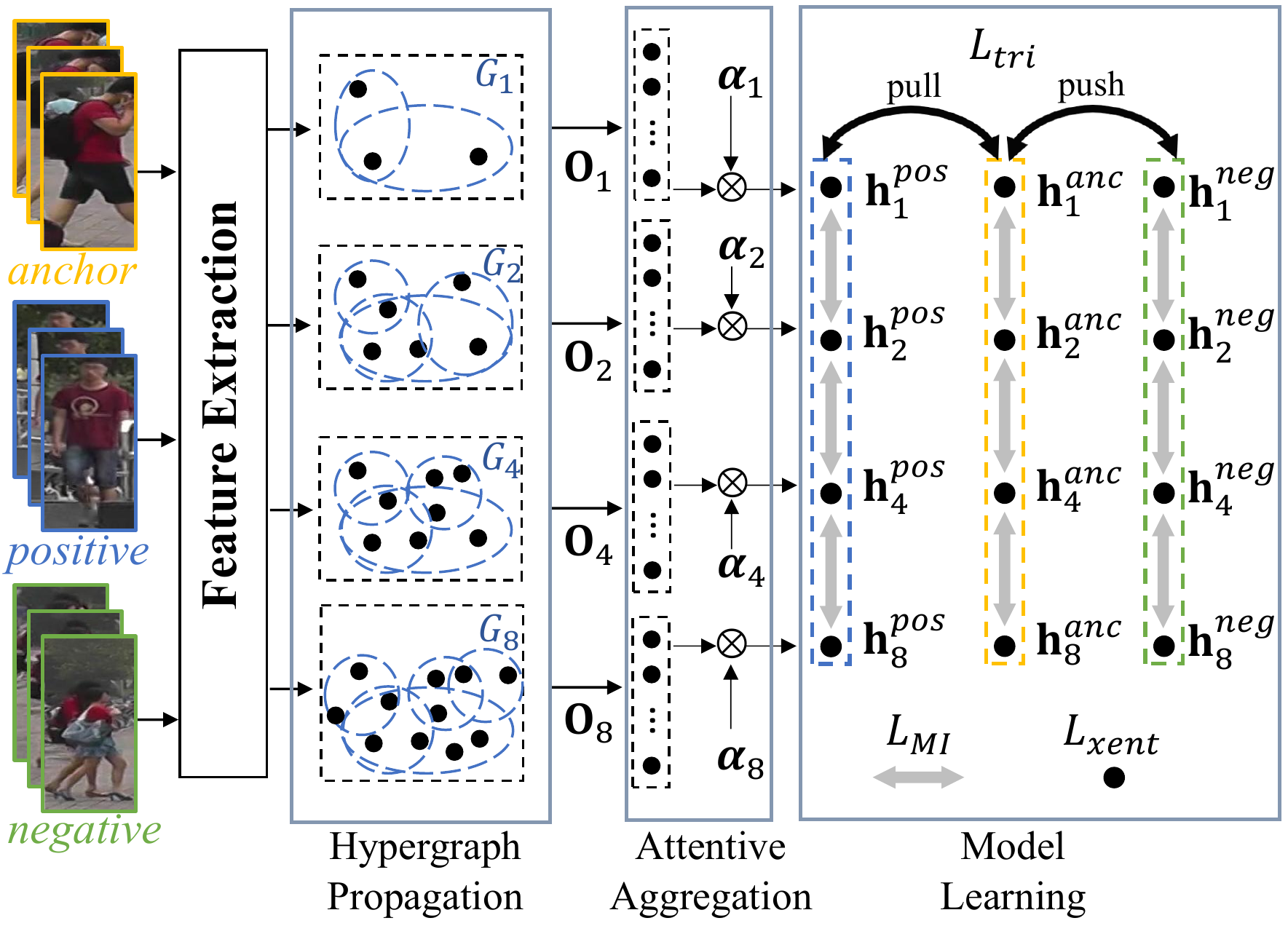}
    \caption{Illustration of attentive node feature aggregation and model learning modules.}
    \label{fig:att}
\end{figure}

\textbf{Attentive Hypergraph Feature Aggregation.}
After obtaining the final updated node features of each hypergraph w.r.t. each spatial granularity, we further need to aggregate node/part-level features into graph/video-level representations for each hypergraph. When deriving aggregation schemes, we should take into account that, within a hypergraph, different nodes are of varying importance. For instance, the occluded parts or backgrounds are less important than the human body parts. It is therefore necessary to develop a specific attention mechanism~\cite{DBLP:journals/corr/BahdanauCB14,Lu_2019_CVPR} to address this.
As shown in Figure~\ref{fig:att}, we propose an attention module which generates the node-level attention for each hypergraph, in order to select the most discriminative part-level features. For each hypergraph, we calculate the node attention $\bm{\alpha}_p = \{\alpha_1, ..., \alpha_{N_p}\}$ as follows:
\begin{equation}\label{eq:13}
    {u}_{i} = \mathbf{W}_{u}{\mathbf{h}_{i}^{L}},
\end{equation}
\begin{equation}\label{eq:14}
    %\alpha_{pi} = \frac{{\rm exp}(u_{pi})}{\sum_{k\in \mathcal{V}_p} {\rm exp}(u_{pk})},
   {\alpha}_{i} = \frac{{\rm exp}(u_{i})}{\sum_{j} {\rm exp}(u_{j})},
\end{equation}
where $\mathbf{W}_{u}$ is the weight matrix. The hypergraph features are then calculated as a weighted sum of the node features:
\begin{equation}\label{eq:15}
    {\mathbf{h}}_{p} = \sum_{v_i \in \mathcal{V}_p} {\alpha_{i} {\mathbf{h}}_{i}^{L}}.
\end{equation}

\subsection{Model Learning}
To optimize the framework, we adopt the cross entropy loss and triplet loss to jointly supervise the training procedure. The cross entropy loss is formulated as follows:
\begin{equation}
    L_{xent} = -\sum_{i=1}^{N}{\rm log}\frac{{\rm exp}(\mathbf{W}_{y_i} \mathbf{h}_p^{i} + b_{y_i})}{\sum_{k=1}^{C} {\rm exp}(\mathbf{W}_k \mathbf{h}_p^{i} + b_k)},
\end{equation}
where $y_i$ is the label of feature $\mathbf{h}_p^i$, $N$ is the mini-batch size, and $C$ is the number of classes in the training set.
Given a triplet consisting of the anchor, positive, and negative features, \ie, $\{\mathbf{h}_p^{anc}, \mathbf{h}_p^{pos}, \mathbf{h}_p^{neg}\}_{\{p=1,2,4,8\}}$, the hard triplet loss is calculated as follows:
\begin{equation}
\begin{aligned}
L_{tri}=-\sum_{anc=1}^N [m &+\max_{pos=1...N}\|\mathbf{h}_p^{anc}-\mathbf{h}_{p}^{pos}\|_2\\&-\min_{\substack{neg=1...N}}\|\mathbf{h}_p^{anc}-\mathbf{h}_p^{neg}\|_2]_+ ,
\end{aligned}
\end{equation}
where $m$ denotes the margin. 

After training the model based on the two loss terms above, each hypergraph will output discriminative graph-level features. The last step is to aggregate graph features w.r.t. different spatial granularities to form the final video representation. In practice, we find that directly pooling graph-level features may lead to significant information loss as each hypergraph captures the unique characteristic of the corresponding granularity. Therefore, we should maintain the diversity of different levels of graph features. Inspired by the information theory, we attempt to fulfill this goal by \emph{mutual information minimization}. Specifically, we adopt an additional loss that reduces the mutual information between features from different hypergraphs, thus increasing the discriminability of the final video representation by concatenating all the features. Here we denote $\mathbf{H}_p = \{\mathbf{h}_p^{i}\}_{i=1}^{N_c}$ as the graph-level features with $p$ spatial partitions, where $N_c$ is the number of tracklets in the training set. Following~\cite{DBLP:conf/iclr/HjelmFLGBTB19}, we define the mutual information loss as:
\begin{equation}
    L_{MI} = \sum_{p,q \in \{1,2,4,8\}}^{p \neq q}\mathcal{I}(\mathbf{H}_p,  \mathbf{H}_q),
\end{equation}
where $\mathcal{I}$ measures the mutual information between different hypergraph features. %Please refer to our \textbf{Supplementary Material} for more details.

Finally, as shown in Figure~\ref{fig:att}, the overall loss function is a combination of the above three terms:
\begin{equation}
    L_{all} = L_{xent} + L_{tri} + L_{MI}.
\end{equation}

\section{Experimental Results}
We evaluate MGH on three benchmark datasets, \ie, MARS~\cite{DBLP:conf/eccv/ZhengBSWSWT16}, iLIDS-VID~\cite{DBLP:conf/eccv/WangGZW14}, and PRID-2011~\cite{DBLP:conf/scia/HirzerBRB11}. We first conduct a comprehensive ablation study to verify the contribution of each component of our model, and then compare our model with recent state-of-the-art approaches.

\subsection{Experimental Setup}\label{sec:imp}
\textbf{Datasets.} \textbf{MARS}~\cite{DBLP:conf/eccv/ZhengBSWSWT16} is one of the largest public datasets for video-based person re-ID, which consists of 1,261 pedestrians captured by six cameras, and each individual appears in at least two cameras. Meanwhile, each identity has 13.2 tracklets on average.
%This dataset was captured in a university campus.
\textbf{iLIDS-VID}~\cite{DBLP:conf/eccv/WangGZW14} contains 600 image sequences of 300 people from two non-overlapping camera views in an airport arrival hall. The frame lengths in each sequence vary from 23 to 192, with an average length of 73. \textbf{PRID-2011}~\cite{DBLP:conf/scia/HirzerBRB11} was collected in an uncrowded outdoor environment with a relatively clean background. This dataset includes 749 people from two camera views, but only the first 200 are captured by both cameras. The length of sequence varies from 5 to 675, with an average of 100. Following previous practice~\cite{DBLP:conf/eccv/WangGZW14}, we only utilize the sequence pairs with more than 21 frames.

\textbf{Evaluation Protocols.} In terms of MARS, we use the predefined training/test split, \ie, 8,298 sequences of 625 people are used for training, and 12,180 sequences of 636 people are used for testing. As for iLIDS-VID and PRID-2011, we follow the standard evaluation protocol~\cite{DBLP:conf/eccv/WangGZW14}. People are randomly split into two subsets with equal size as training and test sets, and the performance is reported as the average results of ten trials. For all the datasets, we use Cumulative Matching Characteristics (CMC) and mean Average Precision (mAP) to measure the performance.

\textbf{Implementation Details.} We employ ResNet50~\cite{DBLP:conf/cvpr/HeZRS16} pretrained on ImageNet~\cite{DBLP:conf/cvpr/DengDSLL009} as the backbone. Following~\cite{DBLP:conf/cvpr/HouMCGSC19}, we insert non-local blocks~\cite{DBLP:conf/cvpr/0004GGH18} into the network and we resize the input images to 256$\times$128. For each batch, we randomly sample 32 sub-sequences from 8 persons. In practice, we set the sub-sequence length $T=8$, the hypergraph layer $L=2$, and the number of neighbors $K=3$. The default spatial partitions are $(1, 2, 4, 8)$, and the temporal thresholds are $(1, 3, 5)$. The influences of these hyperparameters are analyzed in Section~\ref{sec:sensitivity}. We adopt the Adam~\cite{DBLP:journals/corr/KingmaB14} optimizer with weight decay 0.0005. The initial learning rate is set to 0.0003 and is reduced by a factor of 10 every 100 epochs, with the training stage terminating at the 300-th epoch. We concatenate the graph-level features and use the cosine similarity as the distance metric for matching the final video representations. All the experiments are implemented in PyTorch~\cite{paszke2017automatic}, with a Tesla V100 GPU. 

\begin{figure*}
\setlength{\abovecaptionskip}{-0.1mm}
    \centering
    \includegraphics[width=0.24\linewidth]{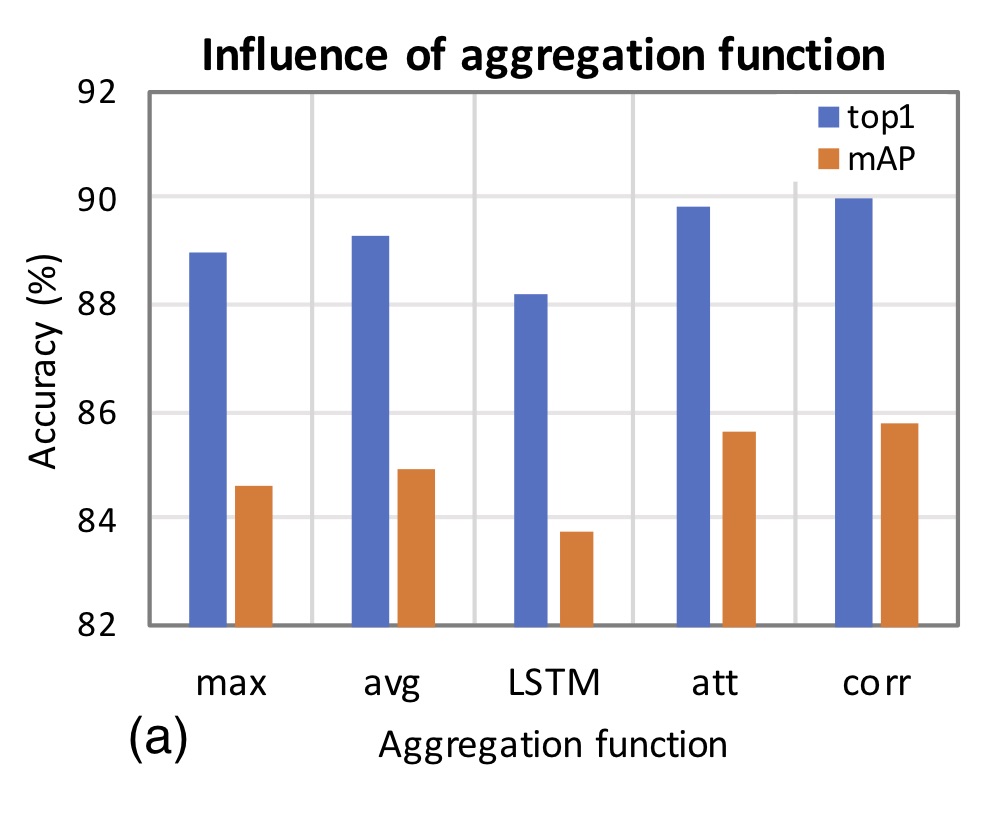}
    \includegraphics[width=0.24\linewidth]{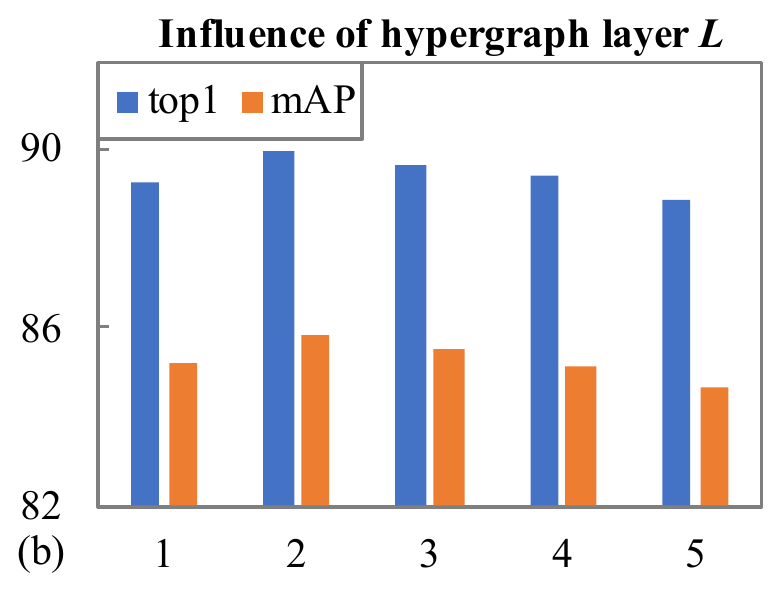}
    \includegraphics[width=0.24\linewidth]{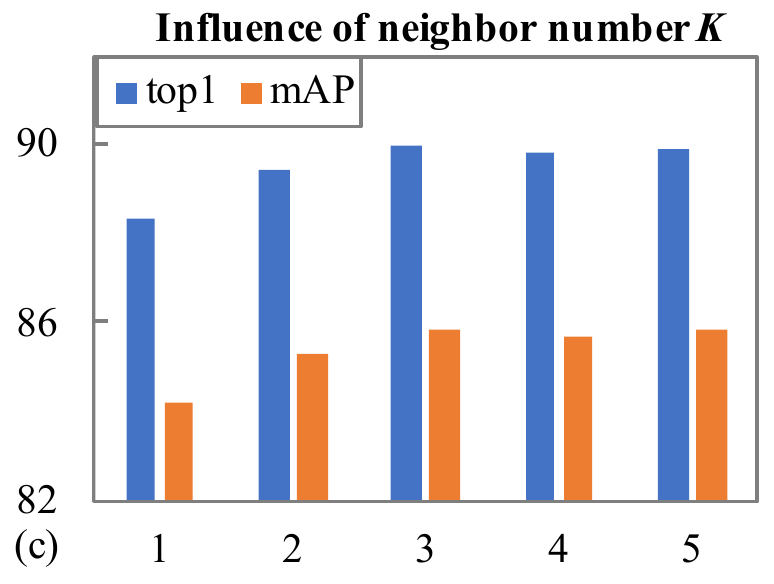}
    \includegraphics[width=0.24\linewidth]{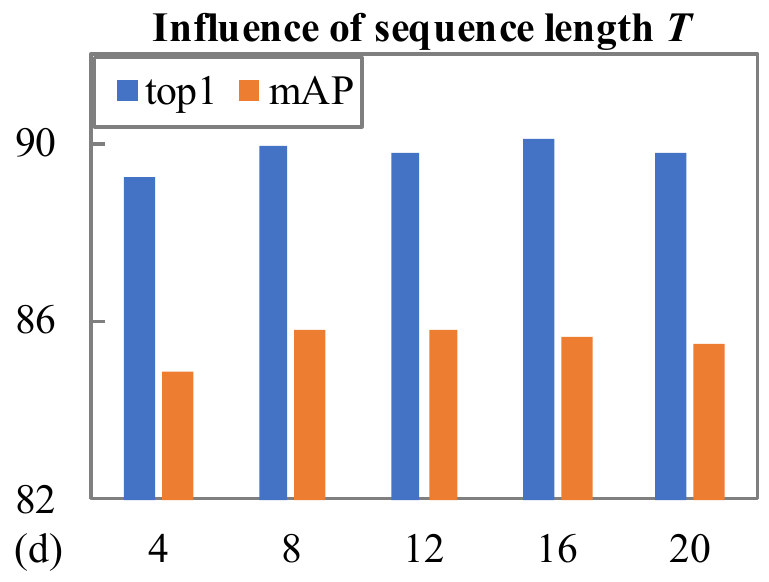}
    \caption{Results on MARS illustrating the influence of different hyperparameter. (a) aggregation function; (b) hypergraph layer $L$; (c) neighbor number $K$; (d) sequence length $T$. Zoom in for best visualization.}
    \label{fig:sensi}
    \vspace{-2mm}
\end{figure*}

\begin{table}[t]
\setlength{\abovecaptionskip}{-0.01mm}
\centering
\begin{tabular}{p{1.7cm}|p{1.1cm}<{\centering}p{1.1cm}<{\centering}|p{1.1cm}<{\centering}|p{1.1cm}<{\centering}}
\hline\thickhline
\rowcolor{mygray} 
  & \multicolumn{2}{c|}{MARS} & \multicolumn{1}{c|}{iLIDS} & \multicolumn{1}{c}{PRID}  \\ \cline{2-5} 
\rowcolor{mygray} 
\multirow{-2}{*}{Methods}  & mAP & top-1  & top-1  & top-1  \\  \hline \hline     
baseline
  & 78.3    & 85.7  & 79.7  & 88.6   \\\hline
HGNN  & 84.1   & 88.7       & 84.3   & 93.8    \\ 
+ Att  & 84.8   & 89.2       & 85.1   & 94.2  \\
+ $L_{MI}$  & \textbf{85.8}  & \textbf{90.0}       & \textbf{85.6} & \textbf{94.8} \\ \hline
\end{tabular}
\caption{Component analysis of MGH. `Att' denotes node-level attention, `$L_{MI}$' denotes mutual information loss.}
\label{tab:component}
\end{table}

\subsection{Model Component Analysis}
We evaluate the contribution of each component and report the results in Table~\ref{tab:component}. For the baseline in the first row, the backbone network is trained with cross entropy and triplet losses. The results are obtained by performing average pooling on frame-level features across the whole video sequence. Through multi-granular spatial and temporal dependency learning with an HGNN, we observe that the top-1 accuracy increases by 3\% on MARS, and by 5\% on iLIDS-VID and PRID-2011, as shown in the 2nd row of Table~\ref{tab:component}. We then insert the node attention module for graph-level feature aggregation, % As seen from the 3rd row of Table~\ref{tab:component},
the attention modules further improve the top-1 accuracy by 0.5\%-1\%, respectively. Finally, by further incorporating the mutual information loss, the top-1 accuracy and mAP are respectively improved from 85.7\% and 78.3\% to 90.0\% and 85.8\% on MARS. As for iLIDS-VID and PRID-2011, the improvement of top-1 accuracy is more than 5\%.
In summary, the major improvement of the framework comes from our hypergraph learning mechanism, as it captures the dependencies of both spatial and temporal clues. The attention module and the mutual information loss bring additional improvements by learning more discriminative graph and video-level features.
% The above results have well demonstrated the effectiveness of our hypergraph model, as well as the attention modules.

\subsection{Model Sensitivity Analysis}\label{sec:sensitivity}
\textbf{Multi-Granularity.} The key motivation of this work is to make full use of the multi-granular spatial and temporal clues in a video sequence to learn better representations. Here, we conduct detailed experiments to evaluate the effectiveness of multi-granular representations, the results of which are illustrated in Table~\ref{tab:granularity}. We test different combinations of spatial and temporal granularities. Specifically, when the temporal range is equal to one, only adjacent nodes are connected. We observe that the performance increases steadily when more detailed spatial/temporal granularities are captured. We also find that the performance saturates when using four spatial granularities (\ie, 1, 2, 4, 8) and three temporal granularities (\ie, 1, 3, 5).

\begin{table}[t]
\setlength{\abovecaptionskip}{-0.01mm}
\centering
\begin{tabular}{p{1.5cm}<{\centering} p{1.5cm}<{\centering}|p{1.3cm}<{\centering} p{1.3cm}<{\centering}}
\hline\thickhline
\rowcolor{mygray} 
Spatial  & Temporal  & mAP  & top-1  \\  \hline \hline     
1  & 1  & 82.3  & 87.5   \\
1  & 3  & 82.8  & 87.7   \\
1  & 5  & 82.9  & 87.7   \\
1  & 1,3  & 83.2  & 87.9   \\
1  & 1,3,5  & 83.3  & 88.1   \\\hline
1,2   & 1,3,5  & 84.6   & 89.3    \\ \hline
1,2,4   & 1,3,5       & 85.5   & 89.8  \\\hline
 1,2,4,8   & 1,3,5       & \textbf{85.8}   & \textbf{90.0} \\
1,2,4,8   & 1,3,5,7       & 85.7   & \textbf{90.0} \\ \hline
\end{tabular}
\caption{Performance of MGH on MARS under different granularities of spatial and temporal clues. `Spatial' denotes the number of human body partitions, and `Temporal' denotes the temporal range for dependency calculation.}
\label{tab:granularity}
\end{table}

\begin{table*}[t]
\centering
\setlength{\abovecaptionskip}{-0.1mm}
\begin{tabular}{l|c|cccc|ccc|ccc}
\hline\thickhline
\rowcolor{mygray} 
  &  & \multicolumn{4}{c|}{MARS} & \multicolumn{3}{c|}{iLIDS-VID} & \multicolumn{3}{c}{PRID-2011}  \\ \cline{3-12} 
\rowcolor{mygray} 
\multirow{-2}{*}{Methods} &\multirow{-2}{*}{Source} & mAP & top-1  & top-5 & top-20  & top-1  & top-5   & top-20    & top-1  & top-5 & top-20  \\  \hline \hline
CNN+XQDA~\cite{DBLP:conf/eccv/ZhengBSWSWT16} &ECCV16 & 47.6   &65.3   & 82.0  & 89.0   & 53.0 & 81.4 & 95.1 & 77.3   & 93.5   & 99.3       \\
SeeForest~\cite{DBLP:conf/cvpr/ZhouHWWT17} &CVPR17   & 50.7  & 70.6  & 90.0  & 97.6     & 55.2 & 86.5  & 97.0   & 79.4 &94.4 &99.3    \\ 
ASTPN~\cite{DBLP:conf/iccv/XuCG0CZ17}   &ICCV17    &-   & 44 & 70  & 81     & 62  & 86 & 98  & 77    & 95  & 99  \\
STAN~\cite{DBLP:conf/cvpr/LiB0W18} &CVPR18
& 65.8   & 82.3  & -  & -    & 80.2     & - & -  & 93.2  & -  & -       \\ 
ETAP-Net~\cite{DBLP:conf/cvpr/WuLDYO018} &CVPR18   & 67.4   & 80.8 & 92.1  & 96.1  & -  & -  & -  & - & -  & -      \\ 
Snippet~\cite{DBLP:conf/cvpr/ChenLXYW18} &CVPR18  & 76.1   & 86.3  & 94.7  & 98.2     & 85.4  & 96.7 & 99.5  & 93.0    & 99.3  & 100 \\
STA~\cite{DBLP:conf/aaai/FuWWH19} &AAAI19
& 80.8   & 86.3  & 95.7  & -    & -    & -  & - & -    & -  & -       \\
ADFD~\cite{DBLP:conf/cvpr/ZhaoSJL019} &CVPR19  & 78.2  & 87.0  & 95.4 & \red{\textbf{98.7}}    & \red{\textbf{86.3}}  & \blue{\textbf{97.4}}  & \red{\textbf{99.7}}   & 93.9    & \blue{\textbf{99.5}}  &100  \\
VRSTC~\cite{DBLP:conf/cvpr/HouMCGSC19}   &CVPR19   & \blue{\textbf{82.3}}   & \textcolor{green}{\textbf{88.5}}  & \textcolor{green}{\textbf{96.5}}  & 97.4    & 83.4  & 95.5  & 99.5   & -    & -  &-  \\
COSAM~\cite{Subramaniam_2019_ICCV} &ICCV19
& 79.9   & 84.9  & 95.5  & 97.9    & 79.6     & 95.3 & -  & -  & -  & -       \\ 
GLTR~\cite{DBLP:journals/corr/abs-1908-10049} &ICCV19
& 78.5   & 87.0  & 95.8  & \textcolor{green}{\textbf{98.2}}    & \blue{\textbf{86.0}}     & \red{\textbf{98.0}} & -  & \red{\textbf{95.5}}  & \red{\textbf{100}}  & 100       \\ 
AdaptiveGraph~\cite{DBLP:journals/corr/abs-1909-02240} &arXiv19
& \textcolor{green}{\textbf{81.9}}   & \blue{\textbf{89.5}}  & \blue{\textbf{96.6}}  & 97.8    & 84.5  & 96.7  & 99.5 & \textcolor{green}{\textbf{94.6}}  & 99.1  & 100  \\\hline
\textbf{MGH} & - & \red{\textbf{85.8}}   & \red{\textbf{90.0}}       & \red{\textbf{96.7}}   & \blue{\textbf{98.5}}  & \textcolor{green}{\textbf{85.6}} & \textcolor{green}{\textbf{97.1}} &\blue{\textbf{99.5}}   & \blue{\textbf{94.8}}  & \textcolor{green}{\textbf{99.3}} & 100   \\ \hline
\end{tabular}
\caption{Comparison with the state-of-the-art video-based person re-id methods. The three best scores are indicated in \red{\textbf{red}}, \blue{\textbf{blue}} and \textcolor{green}{\textbf{green}}, respectively.}
\label{tab:cmp1}
\vspace{-4mm}
\end{table*}

\textbf{Node Propagation Scheme.} In MGH, we aggregate the hyperedge features by calculating the correlations with the target node, as depicted by Eqs.~\ref{eq:4}-\ref{eq:6}. Here, we compare our correlation aggregator with several alternative aggregation schemes. It is worth noting that pooling-based aggregators and LSTM have also been widely utilized for feature aggregation in GNNs~\cite{DBLP:conf/nips/HamiltonYL17}. However, as shown in Figure~\ref{fig:sensi}(a), they (``max'', ``avg" and ``LSTM") are less effective than the correlation aggregator since they neglect the dependency of hyperedge features w.r.t. the target node. Besides, graph propagation often adopts attention mechanisms~\cite{DBLP:conf/iclr/VelickovicCCRLB18}, which typically concatenate the inputs and utilize a fully connected layer to generate the attention weights. We observe that such an attention scheme achieves comparable performance with our correlation aggregator, but it requires more computational overhead. Overall, these results demonstrate the effectiveness of the correlation aggregator.

\textbf{Number of Hypergraph Layers $L$.}
We evaluate the influence of different numbers of HGNN layers. From Figure~\ref{fig:sensi}(b), we can see that the proposed framework is not sensitive to different numbers of layers. Specifically, a two-layer network achieves slightly better performance than other settings. This is because a one-layer HGNN has insufficient representational capability, whilst multi-layer HGNNs contains too many parameters that bring difficulties to the training step. Therefore, we employ a two-layer HGNN in our framework.

\textbf{Number of Neighbors $K$.}
This hyperparameter controls the number of nodes within a hyperedge. Specifically, when $K=1$, an edge only connects two graph nodes and the hypergraph degrades into a standard graph. As shown in Figure~\ref{fig:sensi}(c), in the beginning, the performance increases as $K$ becomes larger, since more context information is included in the hyperedge. However, the performance becomes saturated when $K>3$. These results validate the effectiveness of employing a hypergraph rather than a standard graph. 
%In our framework, we simply employ $K=3$ to get the best performance with relatively low computational complexity.

\textbf{Sequence Length $T$}.
Last but not least, we train and test the framework with various sequence lengths $T$, and the results are illustrated in Figure~\ref{fig:sensi}(d). Overall, the proposed framework is robust to variations in $T$. We also find that longer sequences generate slightly better performance, since the model can capture wider ranges of temporal dependencies. Meanwhile, employing longer sequences increases the model complexity. Overall, $T=8$ gives the best trade-off between performance and complexity. 

In summary, the proposed MHG is not sensitive to most hyperparameters in the framework. The granularity of the spatial and temporal clues plays a key role in the overall performance, which aligns well with the motivation of the proposed framework.

\begin{figure*}[t]
\setlength{\abovecaptionskip}{-0.01mm}
    \centering
    \includegraphics[width=0.33\linewidth]{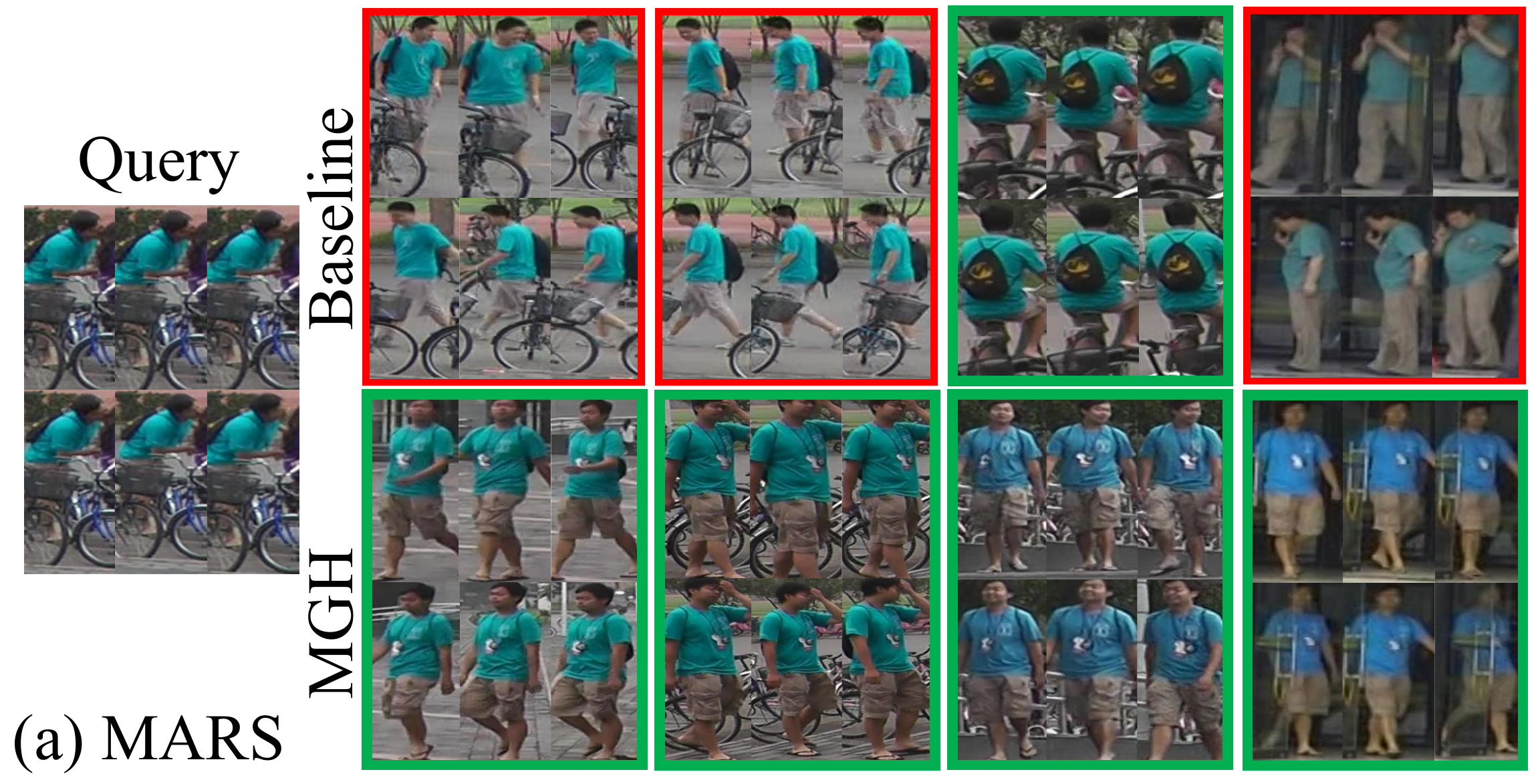}
    \includegraphics[width=0.33\linewidth]{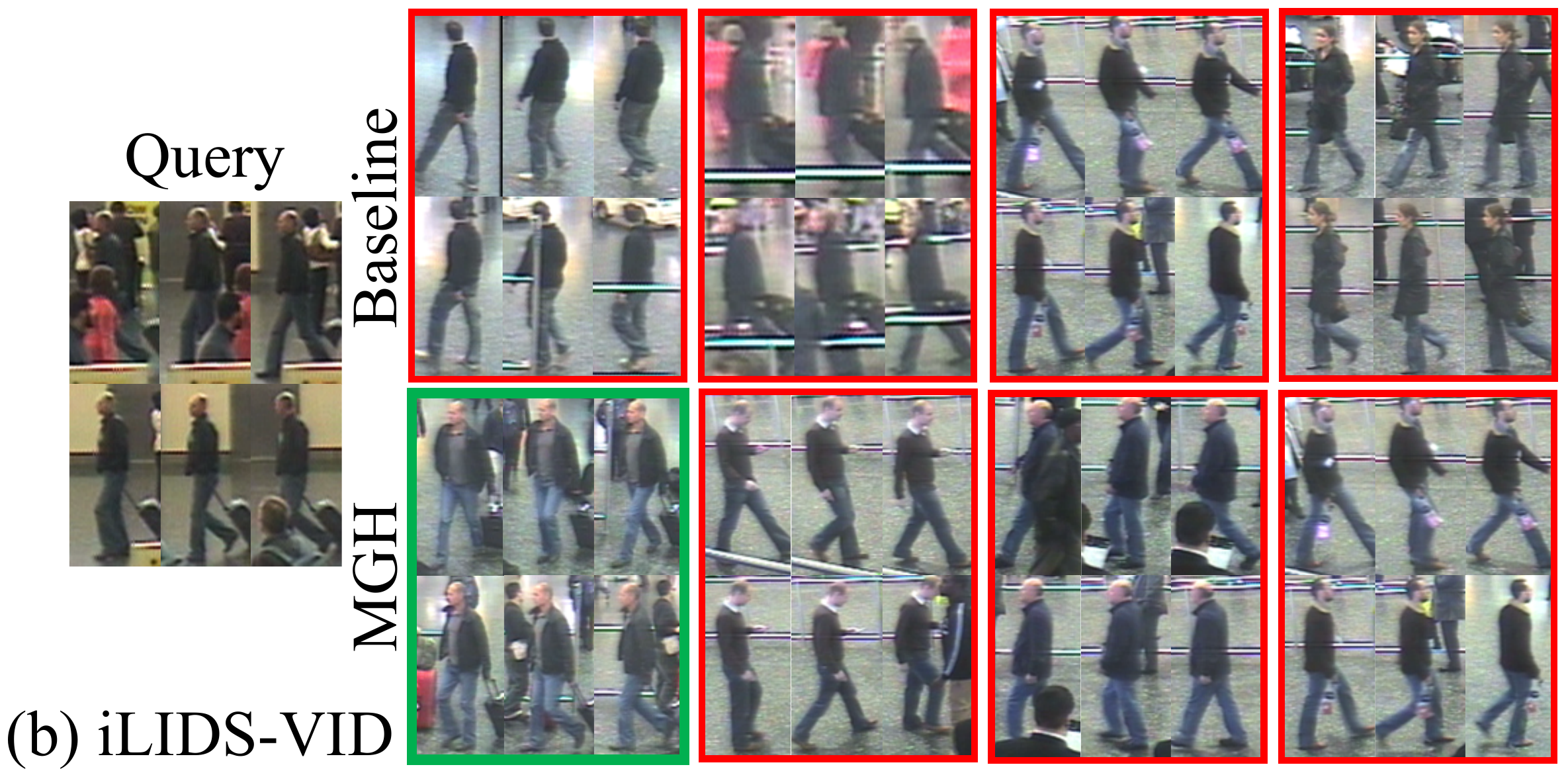}
    \includegraphics[width=0.33\linewidth]{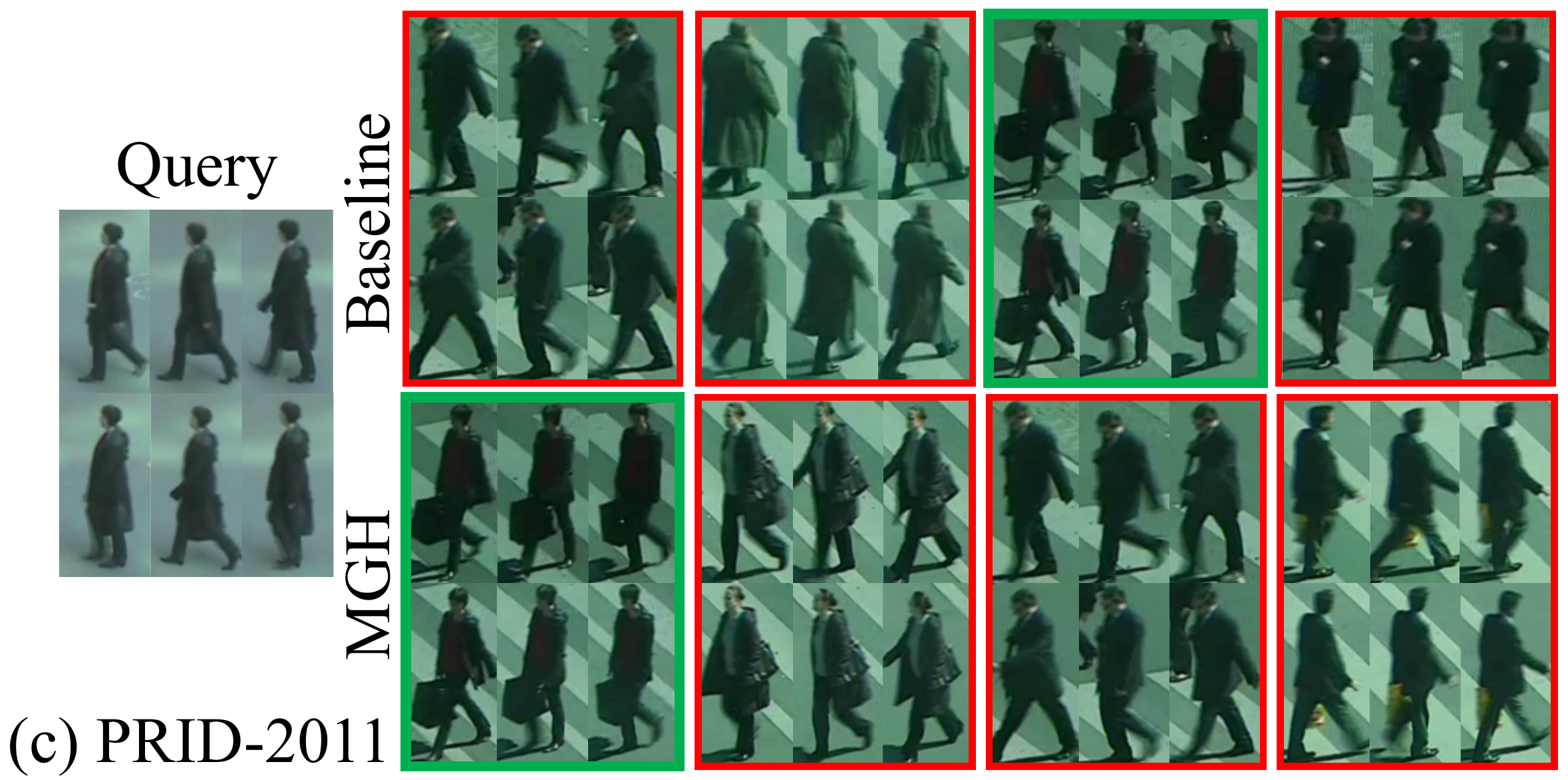}
    \caption{Visualization of re-ID results using the baseline model and the proposed MGH model. The first sequence is the query, whilst the rest are the Rank-1 to Rank-4 (from left to right) retrieved results. The green and red bounding boxes denote correct and incorrect matches, respectively.} %Please refer to our \textbf{supplementary material} for more results.}
    \label{fig:ranking}
    \vspace{-4mm}
\end{figure*}

%\begin{figure}[t]
%    \centering
%    \subfloat[Misalignment\label{fig:att_vis1}]{
%    \includegraphics[width=\linewidth]{fig/vis_att1.pdf}}
%    \hfill
%    \vspace{-3mm}
%    \subfloat[Occlusion\label{fig:att_vis2}]{
%    \includegraphics[width=\linewidth]{fig/vis_att2.pdf}}
%    \vspace{-2mm}
%    \caption{Visualization of node attention weights. The node attention wights are visualized below the input sequences.}
%    \label{fig:vis_att}
%\end{figure}

\begin{figure}[t]
\setlength{\abovecaptionskip}{-0.01mm}
\centering
\includegraphics[width=\linewidth]{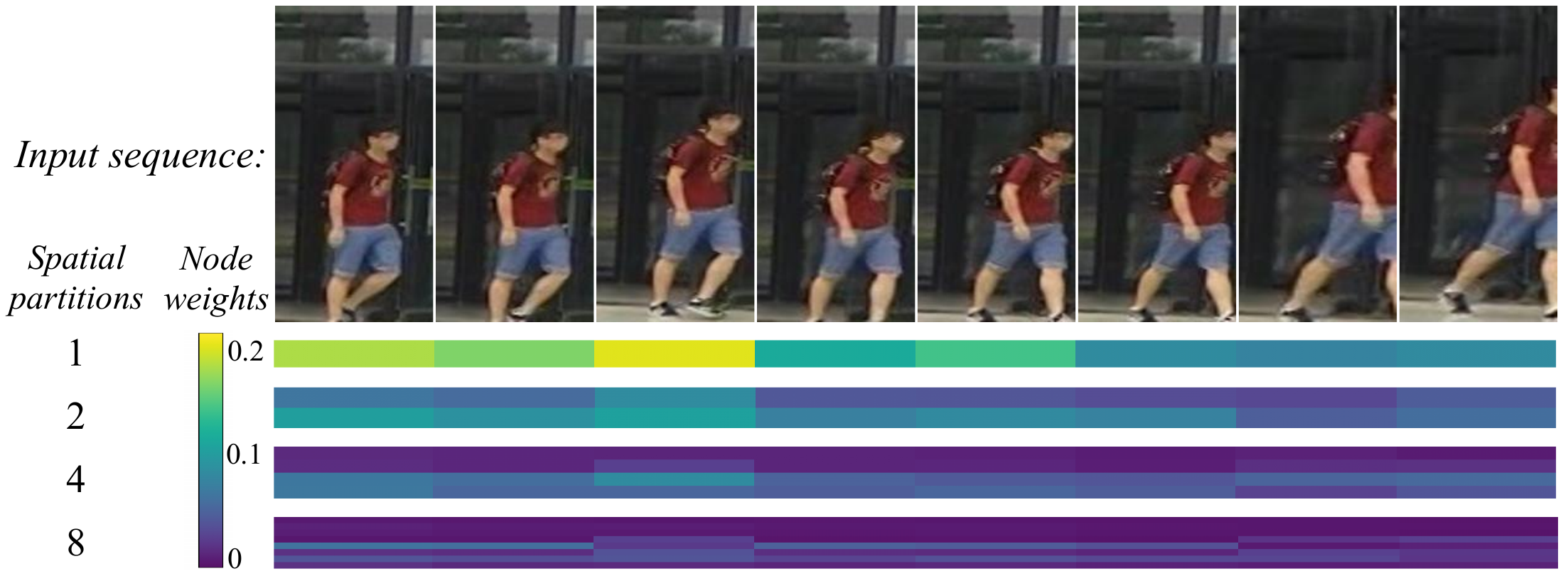}
\includegraphics[width=\linewidth]{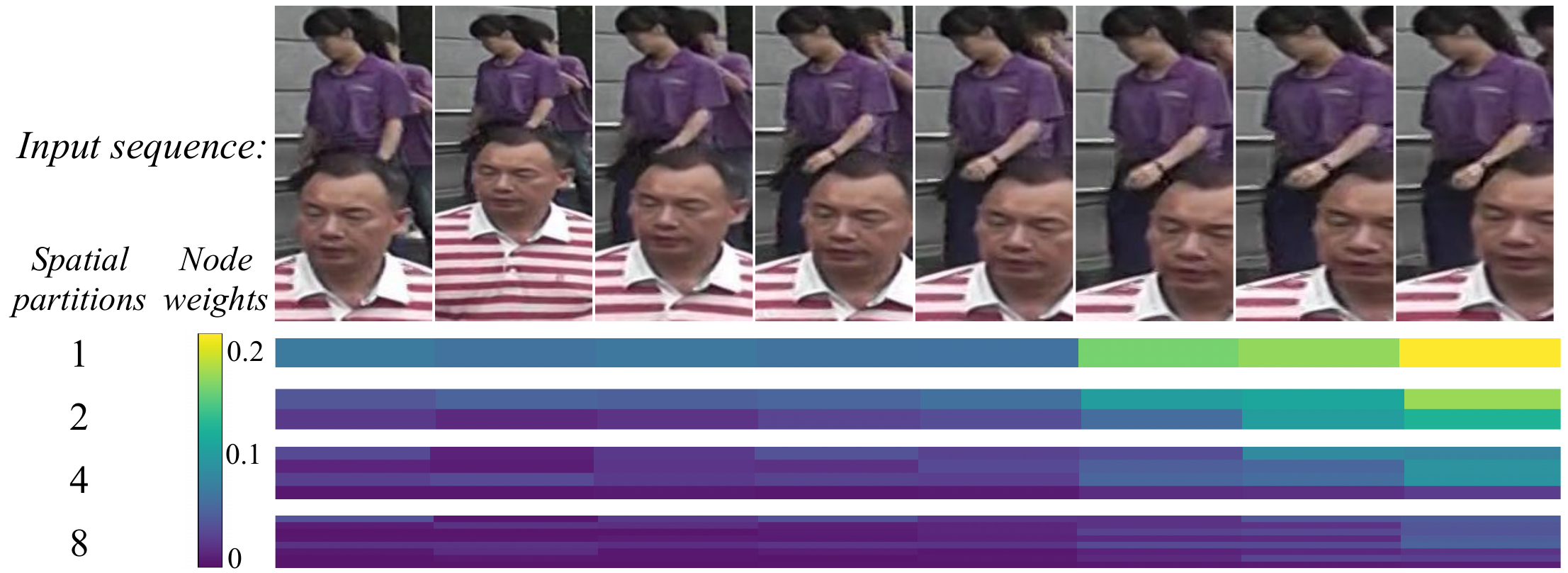}
\caption{Visualization of node attention weights. The node attention wights are visualized below the input sequences.}
\label{fig:vis_att}
\end{figure}

\subsection{Comparisons with State-of-the-Arts}
In this section, we compare the proposed MGH with the state-of-the-art methods on three video-based person re-id benchmarks. The results are reported in Table~\ref{tab:cmp1}. %based on which we have the following observations.

On MARS, our approach achieves 90\% top-1 accuracy, outperforming all previous methods. More remarkably, the proposed model achieves 85.8\% in mAP without re-ranking, showing significant improvement (\ie, 3.5\% higher) over the current best state-of-the-art method. We notice that the recently proposed AdaptiveGraph~\cite{DBLP:journals/corr/abs-1909-02240} also employs GNNs to address the video-based person re-ID task. Our model has two advantages. First, AdaptiveGraph requires additional pose information to build the graph, while our MGH is dynamically constructed based on feature affinities, which makes the proposed model more flexible. Second, AdaptiveGraph only considers the correlation between adjacent frames, neglecting the long-term temporal dependency. In the proposed hypergraph, dependencies in varying temporal ranges are modeled by different hyperedges, yielding more robust representations. It is worth noting that VRSTC~\cite{DBLP:conf/cvpr/HouMCGSC19} also achieves promising results on MARS; however, it is a two-stage model, \ie, VRSTC first locates the occluded regions and completes the regions with a generative model, and then utilizes the non-occluded frames for re-ID. In contrast, the proposed MGH does not need such pre-processing and can be learned end-to-end.

In terms of iLIDS-VID and PRID, since they only contain a single correct match in the gallery set, we only report the cumulative re-ID accuracy. Overall, the proposed MGH achieves competitive results compared with other state-of-the-art methods. Specifically, MGH outperforms several recent models (\ie, SeeForest~\cite{DBLP:conf/cvpr/ZhouHWWT17}, ASTPN~\cite{DBLP:conf/iccv/XuCG0CZ17}, STAN~\cite{DBLP:conf/cvpr/LiB0W18}, ETAP-Net~\cite{DBLP:conf/cvpr/WuLDYO018},
Snippet~\cite{DBLP:conf/cvpr/ChenLXYW18} with optical flow inputs, STA~\cite{DBLP:conf/aaai/FuWWH19} and COSAM~\cite{Subramaniam_2019_ICCV}) in terms of all the evaluation metrics on the two datasets. We note that ADFD~\cite{DBLP:conf/cvpr/ZhaoSJL019} obtains strong performance on iLIDS-VID and PRID-2011. This may be because ADFD employs external attribute labels to learn disentangled feature representations, and such additional information is more effective on small-scale datasets. In contrast, our model only requires identity annotation, and can achieve competitive performance. GLTR~\cite{DBLP:journals/corr/abs-1908-10049} employs dilated temporal convolutions to capture the multi-granular temporal dependencies, and achieves impressive results on these two datasets. However, GLTR does not consider spatial multi-granularity. This is why GLTR obtains less impressive results on MARS, where there tend to be misalignment in the sequence.

In summary, the above results have demonstrated the advantages of the proposed MGH model for video-based person re-ID. With only identity labels, MGH can achieve state-of-the-art performance on MARS, one of the largest existing public benchmarks for this task, in terms of mAP and top-1 accuracy. Meanwhile, MGH also achieves competitive results on iLIDS-VID and PRID-2011.

\subsection{Results Visualization}
We visualize some person re-ID results in Figure~\ref{fig:ranking}. As can be observed, it is difficult for the baseline model to distinguish people sharing similar appearances when there are misalignments and occlusions, resulting in relatively low top-1 accuracy. In these cases, the proposed MGH reduces the visual ambiguity by employing multi-granular spatial and temporal clues. At the same time, MGH achieves more robust results under different illumination conditions.

To better understand the attentive node feature aggregation module, the attention weights of two occlusion examples are visualized in Figure~\ref{fig:vis_att}. On the one hand, for the global spatial partition, the images that contain a larger foreground person tend to be assigned with higher weights. On the other hand, for the local partitions, the parts belonging to the target person have obviously higher weights than the backgrounds. This indicates that the node attention module can adaptively concentrate on the discriminative parts, which validates the effectiveness of the attention module.

\section{Conclusion}
This paper proposed a multi-granular hypergraph learning framework to address video-based person re-ID. The proposed framework explicitly leveraged multi-granular spatial and temporal clues in the video sequence by learning a sophisticatedly-designed hypergraph neural network. In the learning process, we developed an attention mechanism to aggregate node-level features to yield more discriminative graph representations. In addition, we learned more diversified multi-granular features based on a novel mutual information loss. Extensive experiments were conducted on three person re-ID benchmarks, where the proposed framework achieved favorable performance compared with recent state-of-the-art methods.

{\small
\bibliographystyle{ieee_fullname}
\bibliography{egbib}
}

\end{document}